\documentclass[conference]{IEEEtran}
\usepackage{times}
\usepackage{url}
\usepackage[normalem]{ulem}
\usepackage{color}
\usepackage{latexsym}
\usepackage{amsmath}
\usepackage{amssymb}
\usepackage{tabularx}
\usepackage{tabulary}
\usepackage{float}
\usepackage{hyperref}
\usepackage{graphicx}
\usepackage[square,sort,semicolon]{natbib}

\title{A Neural Architecture Mimicking Humans End-to-End \\ for
Natural Language Inference}

\author{\IEEEauthorblockN{Biswajit Paria}
\IEEEauthorblockA{IIT Kharagpur\\
biswajitsc@iitkgp.ac.in}
\and
\IEEEauthorblockN{Annervaz K M, Ambedkar Dukkipati}
\IEEEauthorblockA{Indian Institute of Science, Bangalore\\
\{annervaz.km,ad\}@csa.iisc.ernet.in}
\and
\IEEEauthorblockN{Ankush Chatterjee}
\IEEEauthorblockA{IIT Kharagpur\\
ankushchatterjee@iitkgp.ac.in}
\and
\IEEEauthorblockN{Sanjay Podder}
\IEEEauthorblockA{Accenture Technology Labs\\
\ sanjay.podder@accenture.com}}

\date{}

\begin{document}
\maketitle
\begin{abstract}

%Natural language inference is the problem of identifying whether a
%hypothesis can be inferred or contradicted given a premise, all in
%natural language.
In this work we use the recent advances in representation learning to
propose a neural architecture for the problem of natural
language inference. Our approach is aligned to mimic how a human does the natural
language inference process given two statements. The model uses
variants of Long Short Term Memory (LSTM), attention mechanism and
composable neural networks, to carry out the task. Each part of our
model can be mapped to a clear functionality humans do for carrying
out the overall task of natural language inference. The model is
end-to-end differentiable enabling training by stochastic gradient
descent. On Stanford Natural Language Inference(SNLI) dataset, the
proposed model achieves better accuracy numbers than all published
models in literature.
%and Sentences Involving Compositional Knowledge(SICK) dataset.%
\end{abstract}

\section{Introduction and Motivation}
    The problem of Natural Language Inference (NLI) is to identify
    whether a statement (hypothesis: $\mathcal{H}$) in natural language can be
    inferred or contradicted in the context of another statement
    (premise: $\mathcal{P}$) in natural language. If it can neither be inferred
    nor contradicted, we say hypothesis is `neutral' to premise. NLI
    is one of the most important component for natural language understanding
    systems~\citep{benthem2008brief,maccartney2009extended}. NLI has
    multitude of applications in natural language question
    answering~\citep{harabagiu2006methods}, semantic search, text
    summarization~\citep{lacatusu2006lcc} etc.

    Consider the three statements A: \emph{\texttt{The couple}
    is \texttt{walking} on the \texttt{sea shore}.}
    B: \emph{\texttt{The man and woman} are \texttt{wide awake}.}
    C: \emph{\texttt{The man and woman} are \texttt{shopping on the
    busy street}.} Here the statement A is the premise and, B and C
    both are hypotheses. B can be inferred from A, where as it is
    reasonably clear C cannot be true if A were. A and C can be true
    together, in a strict sense, by arguing that there was a busy
    shopping option by the sea shore, which is not true generally. The
    problem of NLI thus falls in more ``common sense reasoning''
    segment compared to strict logical inferencing and is subtly
    different from deduction in formal logical
    setting~\citep{maccartney2009natural}.

    Unsupervised feature learning and deep
    learning~\citep{bengio2009learning,lecun2015deep} based on neural
    networks have gained prominence in the last few years. State of
    the art neural networks models and appropriate algorithms to train
    these models have been proposed for multitude of tasks in computer
    vision, natural language processing, speech recognition etc and these
    models hold benchmark results for most problems. In
    the area of natural language processing, the recent deep learning
    models have been proven superior to conventional rule-based or
    machine learning approaches in many tasks like part of speech
    tagging, question answering, sentiment analysis, document
    classification~\citep{kumar2015ask} etc. Not only deep learning
    models hold the state of the art results for these problems, many
    model constructs used like \textit{attention mechanism} have close
    alignment with human thought process.

    Motivated by the same, we dissect the problem of NLI into various sub tasks,
    similar to how human carries out NLI. We then realize each sub tasks using a
    deep learning construct, weave them together to create a complete
    end-to-end model for NLI. Let us first see how we can dissect the problem of
    NLI as humans do it. When seeing the two statements A and B as in the example
    above, humans first aligns information snippets between the sentences
    like \texttt{(the couple, the man and woman)} and \texttt{(walking,
    wide awake)}. We notice that first pair is
    equivalent. From the second pair we conclude that walking is
    possible only in the state of being awake. From the results of
    these two different kinds of processing we conclude sentence B
    can be inferred from sentence A. Suppose in A if it were dog
    instead of couple, it would not have been equivalent, we could not
    have inferred B even though the second pair results are the
    same. Each pair results are important, some cases they are
    independent, but in most cases they are dependent as humans make use
    of a lot of contextual information. We analyze shopping on a
    street is not possible at sea shore and conclude C is contradicted
    by A. Note that for inferring B, we never paid attention to where
    the couple were walking, but to contradict C, we paid attention to
    the place. Humans first align the needed information according to
    the context, compare each pair differently by making use of the
    contextual information and then deduce finally by making use of
    each of the comparison results.
    %We propose an end-to-end deep learning model in alignment with this human process.

    The main contributions of this paper are as follows.

\begin{enumerate}
   \item A neural architecture using variants of long short term
   memory, composable neural networks and attention mechanism is
   proposed for the problem of natural language inference.
   \item The model is inspired from how humans carry out the task of
   natural language inference and hence very intuitive. Each step of
   the humans in performing NLI is mimicked by an appropriate deep
   learning construct in the model.
   \item We present detailed experimental results on Stanford Natural
   Language Inference(SNLI) Dataset~\citep{snli:emnlp2015}, and shows that
   proposed model outperforms all the other models
   %and Sentences Involving Compositional Knowledge(SICK)~\citep{marelli2014sick}%
\end{enumerate}

\section{Preliminaries and Background}
    In a deep learning framework, the natural language sentences are converted
    into a numerical representation by word embeddings, in the first place.
    This numerical representations are then encoded by using
    a bi-directional LSTM or a binary tree LSTM, to consider various information
    snippets along with the context in which they appear. Attention mechanism
    is used to learn the parts of the information that needs to be aligned
    and processed together according to the context. The generated pairs
    by attention mechanism are then processed separately using a set
    of different operators selected by soft
    gating. The outputs of the different process pairs are then
    aggregated or composed together for the final prediction
    task. Below we briefly describe concepts of word embeddings and
    LSTMs. Attention mechanism and composition, and their motivations
    are introduced along with the model.
\subsection{Word Embeddings}
    The first challenge encountered in applying deep learning
    models for NLP is to find a correct numerical representation for
    words. ``You shall know a word by the company it keeps''
    (Firth, J. R. 1957:11), is one of the most influential ideas in
    natural language processing. Multiple models for representing a
    word as a numerical vector, based on the context it appears, stem
    from this idea. Many vector representations for words have been proposed,
    including the well known latent semantic
    indexing~\citep{dumais2004latent}. Vector representations for words
    in the context of neural networks was proposed
    in~\citep{bengio2003neural}

    In this paper, each word in the vocabulary is assigned a
    distributed word feature vector, $ w \in \mathcal{R}^m$. The
    probability distribution of word sequences, $P(w_t |
    w_{t-(n-1)},\ldots,w_{t-1})$, is then expressed in terms of these
    word feature vectors. The word feature vectors and
    parameters of the probability function (a neural network) are
    learned together by training a suitable feed-forward neural
    network to maximize the log-likelihood of the text corpora,
    considering each text snippet of fixed window size as a training
    sample.~\citep{mikolov2013efficient}
    adapted this model and
    proposed two new models:  Continuous bag of words and
    skip-gram model, popularly known as `Word2Vec' models. Continuous
    bag of word models try to predict
   % (by maximizing the formed
   % probability measure)
    the
    current word given the previous and next surrounding
    words, discarding the word order, in a fixed context
    window. Skip-gram model tries to predict the surrounding words
    given the current word. These models have better training
    complexity, and thus can be used for training on large corpus. The
    vectors generated by these models on large corpus have shown to
    capture subtle semantic relationships between words, by simple
    vector operations on them~\citep{mikolov2013linguistic}. The
    drawback of these models is that they mostly use local information
    (words in a contextual window). To effectively utilize the
    aggregated global information from the corpus without incurring
    high computational cost, `GloVe'  word vectors were proposed
    by~\citep{pennington2014glove}. This
    model tries to create word vectors such that dot product of two
    vectors will closely resemble the co-occurrence statistics of the
    corresponding words in the full corpus. The model have shown to be
    more effective compared to \textit{Word2Vec} models for capturing
    semantic regularities on
    smaller corpus.
\subsection{Recurrent Neural Network and Long Short Term Memory(LSTM)}
    The basic idea behind Recurrent Neural Networks (RNN) is to capture
    and encode the information present in a given sequence like
    text. Given a sequence of words, a numerical representation (GloVe
    or Word2Vec vectors) for a word is fed to a neural network and the
    output is computed. While computing the output for the next word,
    the output from the previous word (or time step) is also
    considered. RNNs are called recurrent because they perform the
    same computation for every element of a sequence using the output
    from previous computations. At any step RNN performs the following
    computation,
\begin{equation*}
\mathrm{RNN}(t_i) = f(W*x_{t_i} + U*\mathrm{RNN}(t_{i-1})),
\end{equation*}
where $W$ and $U$ are the trainable parameters of the model, and $f$ is a
nonlinear function. The bias terms are left out here
and have to be added appropriately. $\mathrm{RNN}(t_i)$ is the output at
$i^{th}$ timestep, which can either be utilized as is, or can be fed
again to a parameterized construct such as
softmax~\citep{bishop2006pattern}, depending on the task at hand. The
training is done by formulating a loss objective function based on the
outputs at all timesteps, and trying to minimize the loss. The vanilla
RNNs explained above have difficulty in learning long term
dependencies in the sequence via gradient descent
training~\citep{bengio1994learning}. Also training vanilla RNNs is
shown to be difficult because of vanishing and exploding gradient
problems~\citep{pascanu2013difficulty}. Long short term
memory (LSTM)~\citep{hochreiter1997long}, a variant of RNN is shown to be
effective in capturing long-term dependencies and easier to train
compared to vanilla RNNs. Multiple variants of LSTMs have been proposed
in literature.  One can refer to~\citep{greff2015lstm} for a comprehensive
survey of LSTM variants. %The version we are using is explained next.

     A LSTM module has three parameterized gates, input gate ($i$),
     forget gate ($f$) and output gate ($o$). A gate $g$ operates by
 \begin{equation*}
 g_{t_i} = \sigma(W^g*x_{t_i} + U^g*h_{t_{i-1}}),
 \end{equation*}
 where $W^g$ and $U^g$ are the parameters of the gate $g$, $h_{t-1}$ is
     the hidden state at the previous time step and $\sigma$ stands
     for the sigmoid function. All the three gates have the same
     equation form and inputs, but they have different set of
     parameters. Along with hidden state, LSTM module also has a cell
     state. The updation of the hidden state and cell state at
     anytime step are controlled by the various gates as follows,

\begin{equation*}
C_{t_i} = f_{t_i}*C_{t_{i-1}} + i_{t_i}*\mathrm{tanh}(W^C*x_{t_i}+U^C*h_{t_{i-1}})
\end{equation*}
and
\begin{equation}
h_{t_i} = o_{t_i} * \mathrm{tanh} (C_{t_i}),
\label{eq:lstmcellhiddenstate}
\end{equation}
where $W^C$ and $U^C$ are again parameters of the model. The key
component of the LSTM is the cell state. The LSTM has the capability
to modify and retain the content on the cell state as required by the
task, using the gates and hidden states. While forward LSTM takes the
input sequence as it is, a backward LSTM takes
the input in the reverse order. A backward LSTM is used to capture the
dependencies of a word on future words in the original sequence. A
concatenation of a forward LSTM and a backward LSTM is known as
bi-directional LSTM (bi-LSTM)~\citep{greff2015lstm}.

\subsubsection{Binary Tree Long Short Term Memory}
     The LSTM or bi-LSTM model process the information in a sequential
     manner, as a linear chain. But a natural language sentence have
     more syntactic structure to it, and the information is represented
     more as a tree structure than a linear chain. To incorporate this
     way of processing information, tree structured LSTMs were
     introduced
     by \citeauthor{tai2015improved}~\citeyear{tai2015improved}. In a
     tree structured
     LSTM (Tree-LSTM) each node will have multiple previous
     time steps, one each corresponding to a child in the tree
     structure for the node, compared to a single previous time step
     of a linear chain. Different set of parameters for each child is
     included for the input and output gates to learn how different child
     information have to be processed. Using separate parameters child
     information is summed up to form the input and output gate values
     of every node, as follows:
\begin{equation*}
g_{t_i} = \sigma \left( W^g*x_{t_i} + \sum_{l \in \mathrm{child}(i)} U^g_{t_l}*h_{t_l} \right).
\end{equation*}

     Multiple forget gates (one for
     each child) are included to learn the information from each
     child that needs to be remembered. Forget gate update for each $k\ \in\ \mathrm{child}(i)$ is,
\begin{multline*}
f_{t_i,t_k} = \sigma \left( W^f*x_{t_i} + \sum_{l \in \mathrm{child}(i)} U^f_{t_k,t_l}* h_{t_l} \right).
\end{multline*}
     Then cell state is
     updated based on the forget gate values and cell state of the
     children is below.
\begin{multline*}
C_{t_i} = \sum_{l \in \mathrm{child}(i)} f_{t_i,t_l}*C_{t_l} + \\
            i_{t_i}*\mathrm{tanh}\left(W^C*x_{t_i}+ \sum_{l \in \mathrm{child}(i)} U^C_{t_l}*h_{t_l}\right).
\label{eq:treelstm-cellstate}
\end{multline*}
     Hidden
     states are then computed similar to normal LSTM as given in
     \eqref{eq:lstmcellhiddenstate}. The bias terms are left
     out in all the equations and have to be added appropriately
     wherever needed. All $W$s and $U$s in the above equations are
     model parameters, to be learned.

     The tree structure can be formed by considering the syntactic
     parse of the sentence, leading to different variations of Tree
     LSTM~\citep{tai2015improved}. If we consider the syntactic
     structure, each sample in the training data creates different
     tree structures, leading to difficulty in training the model
     efficiently. To work around this we considered complete binary
     tree structures, formed by pairing adjacent words recursively. We
     call this btree-LSTM in the subsequent discussion.

\section{The Proposed Model}
     The model first encodes the sentences using a normal bi-LSTM or a
     btree-LSTM. This is to consider the different segments of the
     sentence along with the context, which is an essential part of
     human processing as explained earlier. In case of bi-LSTM, the
     encodings are augmented along with the corresponding word vectors
     to create enhanced encodings. In the case of btree-LSTM encodings
     this enhancement is not done since, there is no one-to-one
     correspondence with the number of words in the sentence after the
     encoding. If the bi-LSTM encodings are done there will be $n$
     encodings for a $n$-length sentence, where as if btree-LSTM
     encodings are done, there will be $2n-1$ encodings. btree-LSTMs
     considers more possible phrasal structures(along with the context)
     of the input sentence compared to a bi-LSTM, as shown below.
\begin{equation*}
(v_1, \cdots, v_n) \leftarrow \mathrm{bi-LSTM}(\mathcal{S}),
\end{equation*}
\begin{equation*}
(v_1, \cdots, v_{2n-1}) \leftarrow \mathrm{btree-LSTM}(\mathcal{S}),
\end{equation*}
and
\begin{equation}
\mathcal{S}_{e} \leftarrow (s_1,v_1, \cdots, s_n,v_n).
\label{eq:sentenceencodings}
\end{equation}
    The phrase encodings ($v_i$ or $v_i,\:s_i$, $i=1,\ldots,n$) in
    \eqref{eq:sentenceencodings} represents the various
    information snippets in the sentence $\mathcal{S}$ along with the context
    in which they appear. We do this encodings for both the sentences,
    hypothesis $\mathcal{H}$ and premise $\mathcal{P}$. Next phase is
    to align the information snippets between the hypothesis and premise, as humans do,
    for which one can incorporate neural attention.

\subsection{Attention Mechanism}
    Attention mechanism was introduced in the context of machine
    translation recently~\citep{luong2015effective,bahdanau2014neural},
    where in words or phrases from one language has to be mapped or
    aligned to words or phrases in another language for the purpose of
    translating. We use similar concept to learn this alignment for
    our purpose of NLI. Given two sets of vectors, $a
    = \{a_1,..,a_n\}$ and $b = \{b_1,..,b_n\}$, the attention value (a
    numerical quantity) $v_{ij}$ is associated for each element of the
    first set $a_i$ to each element of the second set $b_j$. Forall $\
    a_i\ \in\ a,\ \mathrm{attend}((b_1,\cdots,b_n), a_i)  =  (v_{i1}, \cdots,
    v_{in})$, where,

\begin{equation*}
v_{ij} =  \frac{(b_j)^T a_i}{\sum_r (b_r)^T a_i}
\end{equation*}
One can see that for all $i$,
   $\sum_j v_{ij}\ =\ 1$.
    After learning, attention values will be high for elements
    that are mapped and low for other elements. For example with bi-LSTM
    or btree-LSTM encoding corresponding to \texttt{the man and woman}
    will have high attention value to the encoding
    corresponding to \texttt{the couple}, and low attention values for other
    snippets, in the context of the given sentences. Given an element
    we can generate the attention values and sum up the elements of
    the second set, using the attention values as weights, to create a
    representation of the information that element is interested in or
    aligned with in the second set. As the attention values are high
    only for aligned encodings the summed up vector from the second
    set will be dominated by the aligned information.

    The phrase encodings of hypothesis $\mathcal{H}$ are aligned with the phrase
    encodings of the premise $\mathcal{P}$ using an attention mode as given
    in \eqref{eq:attentionspecific}. The result of the alignment is
    computed using a weighted sum of the phrase encodings of the premise $\mathcal{P}$,
    using attention values as weights.

\begin{equation*}
\mathrm{Forall}\ {\mathcal{H}_e}^p \in\ \mathcal{H}_e,
\mathrm{attend}(\mathcal{P}_e, {\mathcal{H}_e}^p) =  (a_1, \cdots, a_n),
\end{equation*}
where
\begin{equation*}
a_i = \frac{{{\mathcal{P}_e}^i}^T {\mathcal{H}_e}^p}{\sum_j {{\mathcal{P}_e}^j}^T {\mathcal{H}_e}^p}
\end{equation*}
and
\begin{equation}
t_p = \sum_i a_i \cdot {\mathcal{P}_e}^i
\label{eq:attentionspecific}
\end{equation}

    Now that information snippets are aligned, pairs of
    $(t_p,\:{\mathcal{H}_e}^p)$, they need to be processed. Different
    operators have to be applied based on the pairs and context. All
    the individual results have then to be aggregated to make the final
    decision. We use neural network composition for this purpose.

\subsection{Task Composition}

    Often a large task can be solved by composing the results of
    various different sub tasks, each computed separately. Such
    an approach for Question Answering was introduced
    by~\citep{andreas2016learning}. We
    adapt this approach for our
    purpose here. After learning the alignment of encodings, we need
    to perform different functions or comparisons, depending on the
    kind of inputs and the sentence context, to see whether they
    contribute positively or negatively towards final prediction. In
    our example after aligning the encoding corresponding
    to \texttt{the man and woman} with the encoding for \texttt{the
    couple} the model has to process whether they
    are equivalent. Similarly after aligning \texttt{walking} to
    \texttt{wide awake} the model has to do a different
    kind of processing to verify that wide awake is followed from
    walking. Again if in an example, \texttt{all birds} are aligned
    with \texttt{canary}, model might have to check for a type
    of or subset of relationship. Depending on the type of
    input, the context of the sentence, different functions(operators
    or tasks) have to be applied. The operators also has to learn what
    it is supposed to do. Towards this purpose we introduced $k$ number
    of operators, each is a two layer feed forward neural network,
    with different set of parameters. If $a$ and $b$ are the
    aligned encodings corresponding to two different text snippets,
    they are passed through $k$ different two layer feed forward neural
    network, the outputs of each weighed according to a soft gating
    function as
\begin{multline*}
(g_1,\cdots,g_k) = \mathrm{softmax}(W^T [a,b])\\
\mathcal{O} = \sum_i g_i \cdot \sigma \left((W^i_2)^T * \sigma((W^i_1)^T [a,b] \right).
\end{multline*}
    where $W$ s are model parameters. The soft gating function helps to chose which
    operator has to be chosen to be applied, based on their types and context in the sentence in which they appear. Recall from the example different operators have to be applied to compare \texttt{(the couple, the man and woman)} and \texttt{(walking,wide awake)}. This is realized by soft gating function.

\begin{figure}
\centering
\includegraphics[scale=0.3]{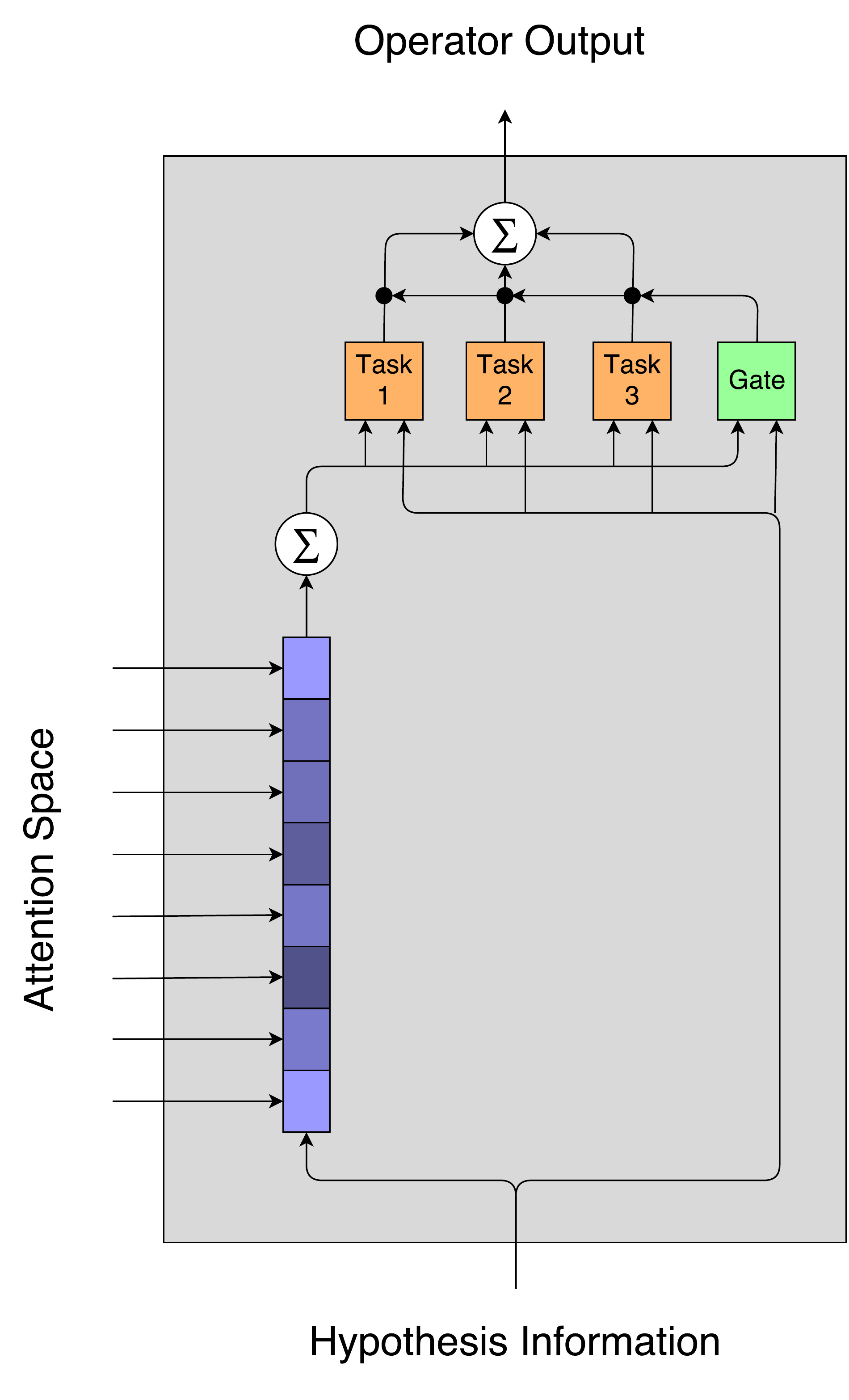}
\caption{Attention and Single Task Module used after bi-LSTM and btree-LSTM encodings}
\label{fig:operator}
\end{figure}

    Expression for $(t_p,{\mathcal{H}_e}^p)$ pairs from
    \eqref{eq:attentionspecific} are given
    in \eqref{eq:gatingspecific}. A schematic diagram of this module is
    given in Figure~\ref{fig:operator}.

\begin{equation*}
\mathrm{task}_i(t_p, {\mathcal{H}_e}^p) =
\sigma\left((W^i_2)^T * \sigma((W^i_1)^T [t_p,{\mathcal{H}_e}^p]\right)
\end{equation*}

\begin{equation*}
(g_1, \cdots, g_k) = \mathrm{softmax}(W^T [t_p, {\mathcal{H}_e}^p])
\end{equation*}
\begin{equation}
\mathcal{O}_p = \sum_i g_i \cdot \mathrm{task}_i (t_p, {\mathcal{H}_e}^p)
\label{eq:gatingspecific}
\end{equation}
Each $\mathcal{O}$ denotes a certain output for an input encoding
pair.  Different pairs yields different $\mathcal{O}$s. All the
$\mathcal{O}$s has to be aggregated(composed) towards the output for
the final prediction. In our example after understanding \texttt{the
man and woman} and \texttt{the couple} are \texttt{equivalent}
and \texttt{wide awake} follows from \texttt{walking}, the
model will have two $\mathcal{O}$ vectors one for each pair. Both the
$\mathcal{O}$s have to be considered in making the final
judgement. How to aggregate the various $\mathcal{O}$ s have to be
learned by the model. There are two parts to it. One is the order in
which they have to be aggregated, if there are more than two. Each
ordering will give a different tree structured computation. The second
being what exactly means aggregation. In the example the aggregation
is an 'and' operator, both has to be satisfied. In another example it
could be 'or' etc. Ideally for this, we should bring in a
reinforcement learning mechanism similar to the one that is used
in~\citep{andreas2016learning}, to learn the order of aggregation and a
neural network~\citep{socher2011parsing} for the learning the aggregation
operator. In our current model(for which results are discussed), we
aggregate the operator outputs $\mathcal{O}$ by using a normal
LSTM. The aggregation order learning which maps to tree structured
computation is envisioned as a part of future work.

\begin{figure*}
\centering
\fbox{\includegraphics[width=160mm, height = 100mm]{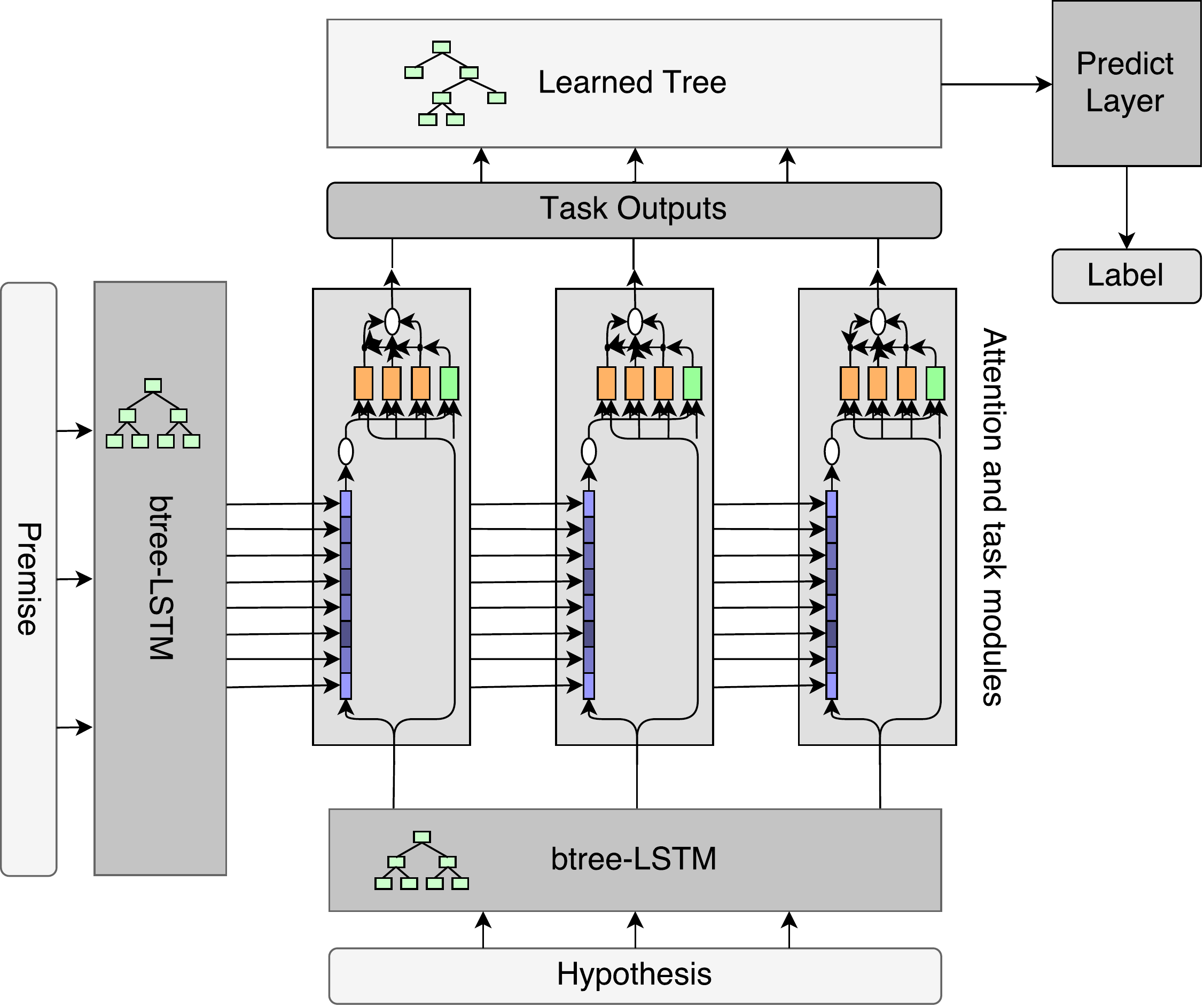}}
\caption{The Complete Model: The upper tree learning(top of the figure) is envisioned for future work, in the current model a simple LSTM is used there instead}
\label{fig:completemodel}
\end{figure*}

The aggregated result $\mathcal{A}$ is then passed through a
comparison layer to do the final prediction, which is shown below,
\begin{equation*}
\mathcal{A} = \mathrm{lstm}(\mathcal{O}_1, \cdots, \mathcal{O}_n),
\end{equation*}
\begin{equation*}
\mathrm{label} = \mathrm{softmax}(W^T \mathcal{A} )),
\end{equation*}
and
\begin{equation}
\mathrm{loss} = H(\mathrm{label}_{\mathrm{gold}}, \mathrm{label}),
\label{eq:loss}
\end{equation}
    where $W$ is the model parameter and $H(p,q)$ denotes the
    cross-entropy between $p$ and $q$. We minimize this loss averaged
    across the training samples, to learn the various model parameters
    using stochastic gradient descent~\citep{stochastic-gradient-tricks}. A schematic diagram of
    the complete model is given in Figure~\ref{fig:completemodel}.

\subsection{Relevant Previous Work}

    NLI is a well studied problem with a rich literature using classical machine learning techniques. With the advent of deep learning, many models including LSTMs were used for NLI. Recently Stanford Natural Language Inference(SNLI) dataset was created~\citep{snli:emnlp2015} using crowd sourcing. Many deep learning models have been benchmarked on this dataset for NLI. Detail list is available at \url{http://nlp.stanford.edu/projects/snli/}. This recent thesis \cite{bowman2016modeling} covers deep learning based works in detail.

    Many of  deep learning based works relied on creating encodings
    of the sentences using LSTMs or convolutional neural
    networks or gated recurrent units
    or variants of recursive neural
    networks, and then using these encodings
    for the final prediction
    task \cite{bowman2016fast,vendrov2015order,mou2016natural}
    are all these kinds of work. \citeauthor{bowman2016fast}
    (\citeyear{bowman2016fast}) also introduced
    an efficient mechanism to learn the binary parse of the tree along
    with creating encodings for the prediction
    task. Works in \cite{rocktaschel2015reasoning,wang2015learning} used
    neural attention mechanism along with LSTMs for the problem of
    NLI.

    There are 3 main works in the space, which claims
    state of the art results. 1. To address the problem of compressing
    a lot of information in a single LSTM cell, \cite{cheng2016long} introduced Long
    Short Term Memory Networks(LSTMN) for Natural Language
    Inference. 2. \citeauthor{munkhdalai2016neural}~(\citeyear{munkhdalai2016neural})
    introduced
    Neural Tree Indexers (NTI), by bringing in attention over tree
    structures of the sentences. 3. \cite{parikh2016decomposable} is
    another very recent work, which uses the
    attention mechanism over words, compare them and then aggregate
    the results. As explained earlier we considers
    attention over possible sentence segment encodings(considering
    context), subtask division, operator selection and
    learning, and aggregation learning. Our model is aligned with human
    thought process and hence very intuitive and achieves state of the art
    results.

\section{Experiments and Evaluation}

    The model was implemented in TensorFlow~\citep{tensorflow2015-whitepaper} - an open-source library for numerical computation for Deep Learning. All experiments were carried on a Dell Precision Tower 7910 server with Nvidia Titan X GPU. The models were trained using the Adam's Optimizer~\citep{kingma2014adam} in a stochastic gradient descent~\citep{stochastic-gradient-tricks} fashion. We used batch normalization~\citep{ioffe2015batch} while training. The various model parameters used are mentioned in Table \ref{tab:modparameters}.

    We experimented with both GloVe vectors trained\footnote{http://nlp.stanford.edu/data/glove.840B.300d.zip}
    on Common Crawl dataset as well as Word2Vec vector trained\footnote{https://code.google.com/archive/p/word2vec/} on Google news dataset. We used Google News trained word2vec word embeddings for the final reported results. Before matching a word in the dataset with a word in the word2vec collection, we converted all characters to lower case. The word embeddings are not trained along with the model. However before using them in our model, we transformed the embeddings using a learnable single layer neural network($\sigma(W^T \cdot w)$, where W is the model parameter and w is the word embeddings). For out of vocabulary words, we assigned them random word vectors. Each element of the word vector is randomly sampled from $\mathcal{N}(0, 0.06)$. This decision has been taken after observing that the word2vec vector elements are approximately distributed according to the above normal distribution.

\begin{table}[!ht]
\centering
\caption{Model \& Training Parameters}
\label{tab:modparameters}
\begin{tabular}{@{}|c|c|@{}}
\hline
\textbf{Parameter Name}        & \textbf{Value}		\\ \hline\hline
Word Vector Dimension & 300        \\ \hline
Sequence Length & 64    \\ \hline
bi-LSTM Hidden State Dimension & 300 \\ \hline
btree-LSTM Hidden State Dimension & 300  \\ \hline
Operator Count & 11  \\ \hline
Batch Size & 40 \\ \hline
Batch Norm $\gamma$ init. value & 0.001 \\ \hline
\end{tabular}
\end{table}

    There are two main datasets available in the public domain for NLI. Sentences Involving Compositional Knowledge (SICK)~\citep{marelli2014sick} dataset from the SemEval-2014 task~\citep{marelli2014semeval} which involves, predicting the degree of relatedness between two sentences, detecting the entailment relation holding between them. SICK consists of 10000 sentence pairs manually labelled for relatedness and entailment. We have experimented with this dataset and have got very good results. The dataset being small and the model having large number of parameters, overfitting could have happened. As benchmark is not available for other state of the art models for comparison on SICK we are not including our results on this dataset. The Stanford Natural Language Inference Corpus(SNLI)~\citep{snli:emnlp2015} dataset contains 570k human-written English sentence pairs manually labeled for balanced classification with the labels entailment, contradiction, and neutral, supporting the task of natural language inference. We are presenting our results on this dataset in comparison with other state of the art models.

\begin{table*}[!ht]
\centering
\caption{Comparison Results on SNLI Dataset}
\label{tab:comparison_results_snli}
\begin{tabular}{@{}|c|c|c|c|@{}}
\hline
\textbf{Model}  & \textbf{Train Accuracy} & \textbf{Test Accuracy} & \textbf{\#Parameters}         \\ \hline\hline
Classifier(hand crafted features)~\citep{snli:emnlp2015} & 99.7 & 78.2 & – \\ \hline
GRU encoders~\citep{vendrov2015order} & 98.8 & 81.4 & 15.0M \\ \hline
Tree-based CNN encoders~\citep{mou2016natural} & 83.3 & 82.1 & 3.5M \\ \hline
SPINN-NP encoders~\citep{bowman2016fast} & 89.2 & 83.2 & 3.7M \\ \hline
LSTM with attention~\citep{rocktaschel2015reasoning} & 85.3 & 83.5 & 252K \\ \hline
mLSTM~\citep{wang2015learning} & 92.0 & 86.1 & 1.9M \\ \hline
LSTM Networks~\citep{cheng2016long} & 88.5 & 86.3 & 3.4M \\ \hline
word-word attention and aggregation~\citep{parikh2016decomposable} & 90.5 & 86.8 & 582K \\ \hline
NTI with global attention~\citep{munkhdalai2016neural} & 88.5 & 87.3 & - \\ \hline\hline
Our model with bi-LSTM encoders & \textbf{89.8} & \textbf{86.4} & 6M\\ \hline
Our model with btree-LSTM encoders & \textbf{88.6} & \textbf{87.6} & 2M\\ \hline
\end{tabular}
\end{table*}

    The comparison results of various models on SNLI dataset is given
    in Table~\ref{tab:comparison_results_snli}. One can see that our
    model with bi-lstm encodings have better accuracy numbers compared
    to all published results, but fall short very close to the results
    reported in not yet published
    works~\citep{parikh2016decomposable,munkhdalai2016neural}. The
    model with btree-lstm encodings have better accuracy numbers than
    all the models. The class
    level accuracy results of various models on SNLI dataset is given
    in Table~\ref{tab:classlevelaccuracy}.

\begin{table*}[!ht]
\centering
\caption{Class Level Accuracy. N: Neutral Class, E: Entailment Class,
  C: Contradiction}
\label{tab:classlevelaccuracy}
\begin{tabular}{@{}|c|c|c|c|@{}}
\hline
\textbf{Method}          & \textbf{N} & \textbf{E} & \textbf{C} \\ \hline \hline
SPINN-NP encoders~\citep{bowman2016fast} & 80.6 & 88.2 & 85.5 \\ \hline
mLSTM~\citep{wang2015learning} & 81.6 & 91.6 & 87.4 \\ \hline
word-word attention and aggregation~\citep{parikh2016decomposable} & 83.7 & 92.1 & 86.7 \\ \hline \hline
Our model with bi-LSTM encoders &\textbf{84.3} & \textbf{90.6} & \textbf{86.9}\\ \hline
Our model with btree-LSTM encoders &\textbf{84.8} & \textbf{93.2} & \textbf{87.4} \\ \hline
\end{tabular}
\end{table*}

\section{Conclusion \& Future Work}

We presented a complete deep learning model for the problem of natural
language inference. The model used deep learning constructs like LSTM
variants, attention mechanism and composable neural networks to mimic
humans for natural language inference. The model is end-to-end differentiable,
enabling training by simple stochastic gradient descent. From the initial experiments,
the model have better accuracy numbers than all the published models. The model is
interpretable in close alignment with human process while performing
NLI, unlike other complicated deep learning models. We hope further
experiments and hyper parameter tuning will improve these results further.

There are different enhancements for the model possible and potential future work
directions. The btree-LSTM currently uses a complete
binary tree structure formed by considering neighbouring encodings
recursively. A binary tree learning scheme, similar to
~\citep{bowman2016fast} can be considered to be incorporated in the
model. Tree construction based on the ordering of attention values
will lead to heap like structures. We are currently working on
this model which we have named \textit{Heap-LSTM}. Currently the model uses
soft gating for operator selection, hard selection with
appropriate learning mechanism is something that has to be
explored. The model aggregates the operator outputs
using a simple LSTM, the aggregation tree structure learning using
appropriate learning mechanism similar to~\citep{andreas2016learning}
is another major stream of work.

The alignment of the model with human thought process, already better accuracy
numbers than all published models just from initial experiments,
all advocate the exploration of model enhancements in these
directions.

\bibliographystyle{plainnat}
\bibliography{NLI}

\end{document}